\documentclass{article}

\usepackage[top=1in, bottom=1in, left=1.3in, right=1.3in]{geometry}

\usepackage[style=apa, backend=biber]{biblatex}
\DeclareLanguageMapping{english}{english-apa}
\usepackage{graphicx}
\usepackage{amssymb}
\usepackage{amsmath}
\usepackage{tcolorbox}
\usepackage{titling}

\title{To be or not to be \\ \large Vector ontologies as a truly formal ontological framework}
\author{Kaspar Rothenfusser}
\date{\today}

\begin{document}

\maketitle

\section{Abstract}

Since Edmund Husserl coined the term "Formal Ontologies" in the early 20th century \parencite{Husserl_2}, a field that identifies itself with this particular branch of sciences has gained increasing attention. Many authors such as \cite{Guizzardi05}, \cite{GFO}, \cite{BFO}, and even Husserl himself \parencite{Husserl_2} have developed what they claim to be formal ontologies. I argue that under close inspection, none of these so claimed \textit{formal ontologies} are truly formal in the Husserlian sense. More concretely, I demonstrate that they violate the two most important notions of formal ontology as developed in Husserl's Logical Investigations, namely a priori validity independent of perception and formalism as the total absence of content. 
I hence propose repositioning the work previously understood as \textit{formal ontology} as the \textit{foundational} ontology it really is. This is to recognize the potential of a  \textit{truly formal ontology} in the Husserlian sense. Specifically, I argue that formal ontology following his conditions allows us to formulate ontological structures, which could capture the fundamentals of what is more objectively without presupposing a particular framework arising from perception. I further argue that the ability to design the formal structure deliberately, without making assumptions about its content, allows us to create highly scalable and interoperable information artifacts. As concrete proof, I showcase that a class of formal ontology, which uses the axioms of vector spaces, is able to express most of the conceptualizations found in foundational ontologies while deducting them as a priori occurrences independent of human perception. Most importantly, I argue that many information systems, specifically artificial intelligence, are most likely already using some type of vector ontologies to represent reality in their internal worldviews and elaborate on the evidence that humans do as well. I hence propose a thorough investigation of the ability of vector ontologies to act as a human-machine interoperable ontological framework that allows us to understand highly sophisticated machines and machines to understand us.

\section{Philosophical Considerations and the Scope of This Work}

I acknowledge that this work intersects with fundamental philosophical questions, such as whether a subject can identify the objective—an issue central to thinkers from Descartes to Kant and Hegel. However, it is neither the objective of this paper to resolve these debates nor do I claim sufficient expertise in the field to do so. While these inquiries are foundational to ontology, this paper does not seek to answer the epistemological problem of subject-object relations but rather proposes a formal framework for structuring ontological representations, particularly in computational systems. The framework of vector ontologies does not claim to identify the objective in an absolute philosophical sense but rather to provide a mathematically rigorous, interpretable structure that allows for scalable, interoperable ontological modeling. This is in contrast to traditional categorical ontologies, which, while formally expressed, derive their structure from recursive abstraction rather than a priori mathematical formality. While dialectical approaches provide valuable insights into dynamic and layered realities, this paper focuses on constructing a rigorous formal structure for analyzing ontological patterns across human and machine representations. Future work may explore how dialectical methods and stratified ontologies could complement this framework.

\section{Edmund Husserl and the Origins of Formal Ontology}

Husserl's \textit{Logische Untersuchungen} (i.e., Logical Investigations) appeared in the early 20th century as an in-depth analysis of the position logic takes within the sciences. His main focus is a structured critique of psychologism, a position held by J.S. Mill, among others, which claims that fundamental laws in logic, and thereby logic itself, are a product of thought, perception, and hence the psyche. Husserl forcefully disagrees with this notion and, in meticulous detail, positions Logic as an \textit{a priori valid} science independent of observation and thought \parencite{Husserl_1}. As such, he advocates for logic being necessarily exact and true by self-evaluation and internal consistency. Thus, he explicitly demarcates it from the probabilistic inductive methods used in practical sciences, including psychology, which he claims concern \textit{"merely empirical, i.e., approximate, laws."}  not truth \parencite[p. 47]{Husserl_1}.
He furthermore argues logic to be a formal science in the mathematical sense, meaning that it is and should be concerned with form and structure, which is independent of its content \parencite{Husserl_1}.
\\
In the continuation of his work, the second volume of Logische Untersuchungen \parencite{Husserl_2}, he coins the term \textit{Formal Ontology}, which, similar to formal logic, should be concerned with "formal concepts and propositions, which lack all 'matter' or 'content'" \parencite[[p. 19]{Husserl_2} and are clearly distinct from the material ontologies which do in fact deal with such content in the form of domain-specific objects predicates and laws \parencite{Husserl_2}.
\\
\\
As such, he contributes two main arguments to the discussion of ontology and formality:
\begin{enumerate}
    \item There are \textit{a priori} valid structures independent of humans and their perception (e.g., Formal Logic), which reach validity through internal consistency.
    \item There lies great value in the investigation of Formal Ontology (from Greek ōn, ont- ‘being’ + -logy 'sciences'), that is, structures concerned with "what is"  in a formal sense without presupposing any content or empirically derived facts.
\end{enumerate}

These contributions are incorporated widely in the philosophy of the following century. A field that identifies itself as Formal Ontology emerges, which is ever-growing in size \parencite{Guizzardi05} and has had a large positive impact on science and philosophy. Even Husserl himself, in the same writing, as he identifies its necessity, starts building what he claims to be such a formal ontology by increasingly abstracting concepts until they lack any meaning and represent "objective categories" \parencite{Husserl_1}. These categories include, among others, "Object, State of Affairs, Unity, Plurality, Number, Relation, Connection, etc." \parencite[p.153]{Husserl_2}, which, as Husserl claims, build a formal ontology as in the above-defined sense. He also investigates what will be established in the field as the key questions of formal ontology, such as "a theory of the pure forms of wholes and parts" \parencite[p.25]{Husserl_2}, also referred to as Mereology. 

What follows is a range of formalized languages and definitions of foundational ontologies developed by various authors \parencite{Guizzardi05, BFO, GFO} that have been positioned as the field of Formal Ontology. This work has been incredibly impactful, specifically in modern Computer Sciences, and in no way is it my intention to question its validity. Quite contrarily, it is that corpus of knowledge that contributes much of the required vocabulary and theory for this argument \parencite{Husserl_1, Husserl_2, Guizzardi05}\footnote{For readers unaware of the work in the field of conceptual modeling and formal ontology I specifically encourage further reading and recommend \cite{Guizzardi05} as a comprehensive overview.}. I also do not argue that they aren't formal in any sense. In fact, most of the literature is written in formal logic and rigorously deducted from its premises. 
However, I argue that while their method is formal, the proposed premises, and hence the produced ontologies, are foundational, not formal. Specifically, I argue that the ontologies created violate the two above-clarified key notions of a Husserlian \textit{formal ontology}, that is, a priori validity and total lack of empirically derived content.
\\
The methodology used to create formal ontologies across the literature  \parencite{BFO, GFO, Guizzardi05} follows Husserl's lead in the identification of ultimate categories of things. Categorization includes distinguishing between that which endures and that which does not \parencite{Guizzardi05, BFO} between properties relations, and quantities \parencite{GFO, Guizzardi05, BFO}, and different types of things \parencite{GFO, BFO, Guizzardi05}. It is evident that these categories are chosen empirically, and even though they are formally expressed in a logical language, they utilize a non-formal method, namely empirical induction, to decide which categorizations are meaningful and how they behave. As such, all ontologies that use categorization demark groups of things by \textit{presumed ontologically meaningful boundaries}. This makes them necessarily \textit{not a priori} but dependent on a presupposed ontological perspective rooted in human perception. In formal terms, we can say that ontologies built in this way \textit{discover} their axioms through empirical evidence of human perceived meaning rather than defining a priori axioms as done in formal logic.

This simultaneously violates both the existence of validity independent of human consciousness, which Husserl imposes on formal sciences, as well as the lack of content.

The latter is evidently the case.  The recursive abstraction from the concrete into overarching categories can asymptotically approach a total absence of content but can never reach it precisely due to its empirical, inductive nature. An ontology built through recursive abstraction from concrete, as done by Husserl, Herrer, Smith, Guizzardi, and others, can hence never be considered a truly formal one as defined in Husserl's Elaboration on logics \parencite{Husserl_1}. Rather, this methodology leads to foundational ontologies, which effectively capture domain overarching structures yet are still concerned with the domain of the human perceived reality and hence not truly formal.

\section{Formal Ontology}

As demonstrated in regards to logic, formalism in the Husserlian sense refers to a priori valid structures, agnostic of their content, valid by internal consistency.  A good example is first-order logic, which defines sets of axioms independent of any meaning.

Accordingly, a Formal ontology, so I argue, should be a set of axioms defined \textit{prior to} and \textit{independent of} any ontological observation but rather tested for a priori validity. As such, formal ontology does not commit to what is but rather provides a \textit{structure} which holds that what is.  Clearly, the choice of those axioms is important. While there are many sets of axioms that are internally consistent and hence a priori valid, it seems reasonable to assume they will not all be of similar ontological expressivity or realism. 

But how can we choose a set of Axioms if not through empirical observation of ontological phenomena? 

I propose to utilize formal structures that are already developed and well-understood. Concretely, I propose to utilize Mathematical structures such as Sets, Groups, and their extensions as Formal ontologies. In fact, much of the foundational Ontology literature explicitly or implicitly uses Set theory and Graph Theory already to study part-whole relationships and other ontological concepts \parencite{Guizzardi05, BFO, GFO}. However, in opposition to previous work, I propose not to build ontologies empirically with Mathematical constructs as tools, but rather to define formal ontologies as Mathematical structures themselves, and afterward map our ontological understanding onto such predefined form, filling it with content to build foundational ontologies inside them. I argue that this would be a truly formal ontological approach in the Husserlian sense and yield immense benefits.

These benefits are of a dual nature. Firstly, it allows us to reveal a more pure and true shape of reality. This is because we have a neutral structure holding our ontological observations, allowing us to analyze ontological structures from the outside perspective. This is equivalent to observing a two-dimensional function in the x-y-plane from a point shifted in the z direction, which allows us to observe its behavior in 2d space in its entirety precisely because the tool of observation is outside of that plane (in 3d space). Accordingly, I argue that a Formal structure with a priori-defined behavior is required to uncover the shape of our perceived ontology and,  hence,  worldview. 

We observe ontological content and its behavior inside a structure, of which we know the fundamental behavior regardless of its content. This is, we can observe the behavior of what is inside what we fully understand. Our understanding of Mathematical structures is not empirical but formal (a priori). Hence, we can fully understand it in a non-empirical sense. Ontological behavior of the world, on the other hand, is necessarily studied empirically and stays material. If the domain ontology's behavior occurs inside a structure we non-empirically understand, however, we have a framework for empiric measurements and a true \textit{formalization} of what is.

In this light, the second benefit of formalism is the opportunity to choose structures that comply with desired requirements through their form, independent of content. More concretely, we can choose, for example, structures that are compatible with most modern machine learning models or with binary calculations. This relates to the "I" in the FAIR (Findable, Accessible, Interoperable, and Reusable) principle of information artifacts, as introduced by \cite{FAIR}, by ensuring interoperability purely in their form before the population of the ontology with content.

Lastly, but of no less importance, the formal ontology, in this sense, ensures the internal consistency of any worldview described in it. As the axioms create a priori consistent and valid structure, one can not build an inconsistent ontology in the formal sense; contrary to that, however, one can build plenty of foundational ontologies inside a formal ontology that are inconsistent with our observation. Making the process of populating Formal ontologies as much an art as a science, more to that in the next sections.
\\
In the next section, I will demonstrate the above-mentioned benefits by proposing a concrete\textit{ Formal Ontology} and discussing its implications. I will also showcase that our proposed formal structure can be used to explain nearly all aspects of foundational ontologies as understood to date while clearly revealing their root in human perception.

\section{Vector Ontologies}

As a concrete instance of a truly formal ontology, I propose what we will call a vector ontology $V_{ont}$ shaped by the following Axioms:

\begin{enumerate}
    \item \textbf{Commutativity of Addition:}
    \[
    \forall\, \mathbf{u}, \mathbf{v} \in V_{ont} \quad \mathbf{u} + \mathbf{v} = \mathbf{v} + \mathbf{u}.
    \]
    
    \item \textbf{Associativity of Addition:}
    \[
    \forall\, \mathbf{u}, \mathbf{v}, \mathbf{w} \in V_{ont}, \quad (\mathbf{u} + \mathbf{v}) + \mathbf{w} = \mathbf{u} + (\mathbf{v} + \mathbf{w}).
    \]
    
    \item \textbf{Existence of an Additive Identity:}
    \[
    \exists\, \mathbf{0} \in V_{ont} \quad \text{such that} \quad \forall\, \mathbf{u} \in V_{ont}, \quad \mathbf{u} + \mathbf{0} = \mathbf{u}.
    \]
    
    \item \textbf{Existence of Additive Inverses:}
    \[
    \forall\, \mathbf{u} \in V_{ont}, \quad \exists\, (-\mathbf{u}) \in V_{ont} \quad \text{such that} \quad \mathbf{u} + (-\mathbf{u}) = \mathbf{0}.
    \]
    
    \item \textbf{Compatibility of Scalar Multiplication with Field Multiplication:}
    \[
    \forall\, a, b \in \mathbb{F},\, \forall\, \mathbf{u} \in V_{ont}, \quad (ab)\mathbf{u} = a(b\mathbf{u}).
    \]
    
    \item \textbf{Identity Element of Scalar Multiplication:}
    \[
    \forall\, \mathbf{u} \in V_{ont}, \quad 1\mathbf{u} = \mathbf{u},
    \]
    where \(1\) is the multiplicative identity in \(\mathbb{F}\).
    
    \item \textbf{Distributivity of Scalar Multiplication with Respect to Vector Addition:}
    \[
    \forall\, a \in \mathbb{F},\, \forall\, \mathbf{u}, \mathbf{v} \in V_{ont}, \quad a(\mathbf{u} + \mathbf{v}) = a\mathbf{u} + a\mathbf{v}.
    \]
    
    \item \textbf{Distributivity of Scalar Multiplication with Respect to Field Addition:}
    \[
    \forall\, a, b \in \mathbb{F},\, \forall\, \mathbf{u} \in V_{ont}, \quad (a+b)\mathbf{u} = a\mathbf{u} + b\mathbf{u}.
    \]
\end{enumerate}

Mathematicians will recognize these axioms as the axioms of a Vector space\footnote{I chose vector spaces rather than more generally and powerfully tensors to slightly reduce the complexity of the argument. However, as I will mention later, it is really tensors that I propose, including topologies governing them, more to that later}. 
\subsection{Motivation for Vector Ontologies}

The reason I chose the axioms of a vector space as a proposed formal ontology is due to my suspicion that they will have the ability to act as an ontological framework compatible with both human and machine intelligence. It is specifically the latter that has been a largely unsolved challenge in the field of computer sciences despite the field's acknowledgment of its potential value. Creating a scalable, ontological framework that both machines and humans can understand is, hence, the key reason vector ontologies are created. That is due to the overwhelming evidence that they are already in use by both humans and machines.
\\

\textbf{Vector Ontologies and Machines}\\

Especially in recent decades, machine learning utilizing artificial neural networks (ANNs) has gained increasing attention in both research and industry. Originating from the effort to mimic the human brain, they perform complex tasks through weighted neural connections trained with large amounts of data. Despite their success in performing various advanced challenges, they are often criticized rightfully for their \textit{black-box} nature. Concretely, the task performed (e.g., classification of images) is completed using so-called \textit{hidden states}, which perform computations that are not fully understood as to their relation to the questions they answer (e.g., is it a cat or a dog).  The intuition we should have about this is that even though they decide correctly, we can not grasp any sort of logic in how they choose to answer.

The tasks performed by ANNs, such as image recognition (one of the most successful applications), clearly are of an \textit{ontological} nature, for example, the judgments required to distinguish between a cat and a dog. This leads to a paradigm where we know that an ANN has learned \textit{some} ontology, but we have no understanding as to which one. I argue, however, that we do very well know the form of that ontology in the Husserlian sense.

All processing inside neural networks is done through mathematical operations belonging to the class of high-dimensional \textit{Vector spaces} \footnote{The mathematicians and machine learning crowd will correct me herein, as this is a simplification. To be more truthful, ANNs perform Tensor operations, not just vector operations, and include nonlinearity due to activation functions. We recognize this and, in later sections, elaborate on the extension of vector ontologies to tensor ontologies, including complex topologies; however, for the sake of argument, I believe you agree that this is more detail than the core issue. Hence, we will make it easier for non-experts to follow the argument by taking the special case of a vector.}. As such, any ontological task performed in the neural network must necessarily be performed inside an ontology, which has the structure of a vector space\footnote{Again, more correctly, tensors in complex non-linear topologies}. 

Thus, building a formal ontology with the axioms of a vector space allows for a better understanding of the inner workings of ANNs and the design of ontologies that are understandable by ANNs. This clearly has huge potential, as it would allow for a machine-human interoperable ontological framework.
\\
\\

\textbf{Vector ontologies and humans}\\
\\
While the compatibility with machines is clearly a benefit of vector ontologies, the question arises, of course, whether they are similarly compatible with human cognition. I will show evidence that humans use vector ontologies in both our neural processing and our consciousness itself. However, to properly investigate \textit{whether} humans use vector ontologies, we need to establish a mapping of our understanding of the ontological universe into the formal structure of vector spaces.

I will, in the following hence, discuss a possible interpretation of ontological concepts inside this structure. It is important to note that some might rightfully remark that such a mapping is inherently introducing bias and ontological perspective, and I fully agree. However, as we are sure of the structure's (vector space's) internal consistency, we can effectively measure the quality of our mapping in the degree to which it creates ontologically consistent behavior. I accordingly encourage constructive criticism and hopefully better mapping suggestions from my peers. I will, however, show that many concepts discussed in the current literature on foundational ontologies can be consistently described in a vector space using my proposed mapping.

It is important to note that this work was developed completely independently of the work of Gardenfors on conceptual spaces \parencite{Gardenfors}, which has only come to my attention in the last couple of days. Gardenfors provides some similar arguments for the utility of \textit{Conceptual spaces} in understanding human and machine perception and calls to build systems utilizing them. I share with Gardenfors the motivation to find a more productive, interoperable, and constructive representation of information that can connect explanatory with constructive power. In fact, structuring information was one of the core motivations that eventually led to this work. I also agree with many of Gardenfors' arguments and am specifically appreciative of some rigorous empirical support for the connection to human perception. However, I argue that there is an important differentiation in my work, namely the truly formal nature, which assumes the structure and form without content a priori, while Gardenfors bases his investigation on empirical observation. Secondly, I define ontology in its completeness as a vector space, resulting in a tool for the investigation of arising phenomena. This clearly differs from Gardenfors' proposal to utilize conceptual spaces as an explanation for how we perceive. Despite this, I agree with many of his conclusions about conceptual spaces and their implications in full. However, they would exceed the scope of this argument, given that our work was developed independently. I will hence discuss Gardenfors rather tangentially regarding the evidence he provides for humans' relation to vector ontologies.

\subsection{Ontological concepts in Vector-Ontologies}

The ontological concepts discussed in foundational ontology divide themselves across three distinct phenomena in a vector ontology. That is in the structure of the space itself, the existence or nonexistence of a particular point in space (the set of existence vectors), and lastly, patterns of existence vectors, which we will call functions of existence. I apologize in advance for using some terms from foundational ontology without proper explanation. I refer to \cite{Guizzardi05} as a prior reading, which provides a comprehensive summary of the terminology discussed. 

\subsubsection{Structure}

We define the structure of a vector ontology using the basis vectors of a Vector space as follows.

A vector ontology $V_{ont_{D}}$ for a given domain $D$ is defined as a vector space spanned by a set of basis vectors $\{x_1, x_2,x_3,...,x_n\}$.These basis vectors represent the set of domain-specific quality dimensions (or properties) in the literature, also referred to as \textit{universals} \parencite{DOLCE, Guizzardi05, Gardenfors, BFO, GFO}, which we will formalize as  $B_{D}$ (Basis for the Domain). We hence, formally define a vector ontology as:
\[
V_{\text{ont}_D} = \text{span} \{ x_1, x_2, \dots, x_n \}, \quad x_i \in B_D
\]
Intuitively, this means that a Vector ontology is defined as a high-dimensional space where each dimension (basis vector) represents a property of the objects inside it. 
The field used to construct the vector space provides the domain of qualitative values (often called quale in the literature \parencite{Guizzardi05}). For example, a vector space over the field $\mathbb{R}$ encodes continuous properties (such as height). If one wishes to represent discrete properties, one might instead employ a finite field (or another appropriate discrete structure) rather than $\mathbb{R}$. Intuitively, this means that, depending on the type of quality we want to encode, we choose an appropriate set of possible values that each quality can take on. For example, when encoding height, we might want to allow for infinitely many possible values with arbitrary precision (mathematically $\mathbb{R}$), while a dimension encoding the number of edges of a shape might only allow for natural numbers, that is positive integers (mathematically $\mathbb{N}$). 

\begin{tcolorbox}[colback=blue!5!white, colframe=blue!75!black, title=Example Vector Ontology for shelves]
Let \(\mathbb{R}\) be the field of real numbers used to represent continuous quantities (or \emph{qualia}). Define the vector ontology for the domain of shelves, denoted by \( V_{\text{ont}_\text{shelves}} \), as the vector space over \(\mathbb{R}\) with basis:

\[
V_{\text{ont}_\text{shelves}} = \text{span} \{ x_1, x_2 \}, \quad x_i \in B_{\text{shelves}} = \{ \text{height}, \text{width} \}.
\]

where:
\begin{itemize}
    \item \( x_1 \) represents the universal vector corresponding to \emph{height},
    \item \( x_2 \) represents the universal vector corresponding to \emph{width}.
\end{itemize}

Thus, for any shelf entity $s \in V_{ont_\text{shelves}}$, there exist unique scalars $a_1, a_2 \in \mathbb{R}$ such that
\[
s = a_1 x_1 + a_2 x_2.
\]

The choice of a basis ensures that every entity in the ontology (here, every shelf) can be uniquely represented in terms of the selected quality dimensions. In our shelf ontology, the two-dimensional representation reflects the assumption that height and width are the only two properties of interest.
\end{tcolorbox}

In summary, the structure of a vector ontology is described as a finite set of basis vectors $\{x_1, x_2,x_3,...,x_n\}$ representing qualities (universals) linked to a field providing their quantities (quale). 

\subsubsection{Existence}

Ontology (from Greek ōn, ont- ‘being’ + -logy 'sciences') is a science of existence, it aims to distinguish that which is from that which is not. Husserl formulates this clearly in his first logical investigation:

\begin{quote}
"Rather we may say that, if it is to be called 'knowledge' in the
narrowest, strictest sense, it requires to be evident, to have the luminous
certainty that what we have acknowledged is, that what we have rejected is
not, ..." (Edmund Husserl, Logical Investigations, p.17)
\end{quote}

Vector ontologies embody this notion in its purest form. In fact, whether something is or is not is the only information we can directly extract from a vector ontology. What I mean by that is that even with the structure of the space defined above, our ontology is empty (it contains only non-instantiated universals). Describing an object is to describe the existence of a point (vector) inside the vector space that instantiates a domain object.
We take as an example the domain of colored shapes described by a vector ontology with basis vectors representing the properties of number\_of\_edges, redness, greenness, and blueness.
\[
B_{\text{colored-shapes}} = \{ \text{number\_of\_edges, redness, greenness, blueness}\}.
\]
In this domain, a blue Rectangle exists if and only if the vector [4, 0, 0, 255] in fact \textit{exists} in the sense that it is contained in the Vector ontology:

\[
\text{Blue Rectangle exists} \implies [4,0,0,255] \in V_{\text{colored-shapes}}.
\]

So, the existence of an object in a domain is given as the truthfulness of the following expression:
\[
v \in V_{\text{domain reality}}.
\]
where v is the vector representing the object of interest. \\

This reveals the key assumption of vector ontology, which is that any object of interest can be fully described as a vector in a vector space with a finite number of basis vectors representing quality dimensions. It is clear that with the increasing complexity of the objects represented, the number of such basis vectors must grow very large to capture all the fine details. However, I argue that it most likely behaves similarly to a Fourier transform, where there is a small set of most important and dominating universals that explain most of the object with an asymptotic resolution increase as additional basis vectors are added. In fact, one of the most common methodologies used to find Fourier transformations is the identification of an orthogonal basis. Which basis vectors hold the most relevant information varies by use case, and hence, the selection of appropriate dimensions to use is contextual and important. 
For instance, in the above example, one might be less interested in the color of the blue Rectangle but rather in its size or spacial position, in which case other basis vectors (size and Euclidean space) should be chosen or added.
\\

This definition of objects as the truth value of $v \in V_{\text{domain reality}}$ shows an important notion of how we use vector ontologies compared to vector spaces themselves. A vector space in its mathematical form contains all possible vectors; as such, it is continuous or densely populated. A vector ontology, however, is interested in both the dense population, describing all theoretically possible objects, as well as the sparse population of the vector space representing reality, which is what we mean to describe in the expression $v \in V_{\text{domain reality}}$, more to that in the next section.
\\

Other Authors have touched upon the fact that universals (in our case, basis vectors) do not always apply to all objects (e.g., ideas can not be more or less "red" the way tables can be). The definition of objects as a finite series of basis vectors clearly incorporates this notion in that basis vectors are often constructed for subsets of the vector space. Hence a set of basis vectors applicable to domain X might not be applicable to domain Y (Domains representing a subset of the vectors in the vector space). Intuitively, we can understand that in the way that properties/basis vectors that do not apply to a given object are scaled by 0, making them effectively disappear. Similarly, suppose a basis vector has constant scaling across all objects of the set (e.g., a set of all rectangles which makes number\_of\_edges = 4 for all). In that case, we can eliminate the number\_of\_edges dimensions as it bears no information in the domain. Even more so, a constant value in a given dimension might be used to define a domain (e.g., one of rectangles). This is often discussed as manifold theory in machine learning, which states that much of our reality, at least locally, lies in low-dimensional subspaces. As such, it is expected that for a clearly bounded domain, a smaller number of basis vectors will be required to describe its ontology, assuming that we are aware of the domain boundary. In line with this, we can say that the truly universal basis vectors are those which can host all ontological objects of reality. However, in a given domain, we are concerned with a small portion of that vector space (a subset of its vectors), which is more effectively described with a small set of local basis vectors representing a specific domain.\footnote{One might logically conclude that we should choose domain boundaries by the set of basis vectors required to describe its objects in full}\\
As such, the often discussed notion of "context"  materialities in our vector ontological conceptualization as a position in the universal property space that leads to local basis vectors modulating global functions.
\\
\\
In summary, all ontological information lies in the existence or non-existence of specific vectors or sets of vectors in a vector ontology and patterns of vectors that are - hence the title of this paper, "To be, or not to be."

\subsubsection{Patterns of Existence}

Now that we have properly defined the structure of our vector ontology and the meaning of a vector's existence, I will demonstrate in the following paragraphs that it is really \textit{ patterns of existence} that the discipline has so far been investigating. \\

Clearly, describing reality as unconnected sparse vectors in a vector space is of little use. Our ontological perception is concerned with patterns and notable phenomena in the distribution of those points. In the following, we will examine some of those patterns and link them to ontological observations from existing literature.
\\
\\
\textbf{Functions of Existence}\\

At this point, it seems appropriate to introduce what I will call \textit{functions of existence} $f_e$. In Mathematical terms, we introduce linear maps. The intuition behind this is that if a function applied across a given range of inputs gives us a set of vectors as the output of which we know that they exist in the vector ontology, we can extract that function as a pattern in the ontology, which unifies the individual points of existence into more complex, unified structures.

\begin{tcolorbox}[colback=blue!5!white, colframe=blue!75!black, title=Example function of existence]
Given a vector ontology for the domain of \textit{humans} $V_{humans}$ with the basis vectors $x_1 = time$, and  $x_2 = body\_wheight$. If we observe that there is a collection of vectors $[50,68],[51,68],...,[60,68]$ that has constant values $x_2$ in a continuous time frame, we can abstract a pattern expressed as the following function:
\[
f_e(t) = 
\begin{cases}
68\,\text{kg} & \text{if } 49 < t < 61, \\
\text{?} & \text{otherwise}.
\end{cases}
\]

Intuitively, knowing that John weighed 68 Kilograms at age 50 and didn't gain or lose weight for the next 10 years is sufficient to reconstruct his weight at any point in that time frame. This allows us to forget the individual vectors and instead remember the function.

It seems to hold for how we remember patterns. For example, a phone number might be remembered as "three times 0" rather than "0, 0, 0" to have an easier recall. This is called \textit{chunking} in the field of psychology.
\end{tcolorbox}

In this matter, our perception of object boundaries seems to be an information compression of many vectors into a function, allowing us to reconstruct them. For example, instead of storing boolean existence values for indefinitely many points in space representing a sphere, we can simply store the equation that describes whether a point is part of the sphere ($x^2+y^2+z^2 \le r^2$). This is conceptually equivalent to human perception, which unifies millions of molecules in a single concept of a ball of a given radius, material, and center.
\\

To conclude, Functions of Existence, which, mathematically speaking, are linear maps, allow us to unify patterns of existence into a single concept that holds the reconstructive information. As such, they represent compressed definitions of subsets of the vectors that in fact exist, unified through some function of unity.

\begin{tcolorbox}[colback=blue!5!white, colframe=blue!75!black, title=Mathematical comment on Functions of Existance]
Mathematical speaking we are introducing Multilinear maps. It seems that when our perception draws the boundary of an \textit{object}, it essentially learns a multilinear Map. This Map maps from $V_{ont}$ into a binary space of existance $V_\exists$. 

    \centering
    \includegraphics[width=0.7\linewidth]{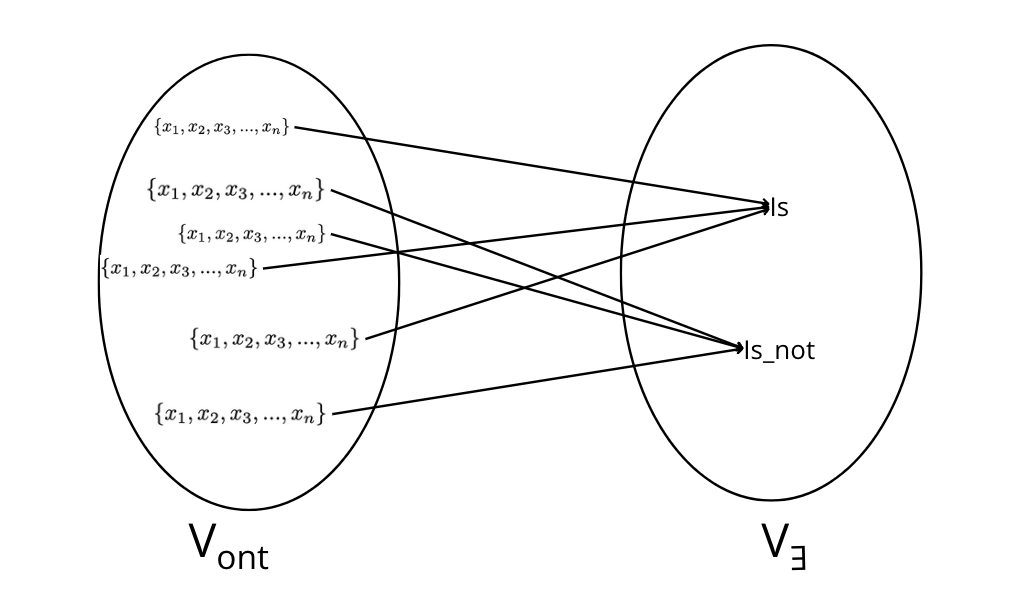}
\\
Computer scientists, specifically when concerned with deep learning, might note already here that it is indeed multilinear maps (matrices) that are learned by Neural networks. This provides additional evidence for the already wide use of the here-called Functions of Existences inside vector structures with the goal of learning concepts.

\end{tcolorbox}

From this, it becomes evident that if we are discussing an object or concept, we are referring to a learned function, which maps a collection of vectors into a conceptual unit. We will later see that Functions of Existence play a crucial role in human perception of parthood and causality. It is, however, noteworthy that these functions and, hence, Maps are not used here by the ontology itself but rather to facilitate our investigation and interpretation of the same.
\\

\textbf{Continuity}\\

A noteworthy aspect of FOEs (functions of existence) is their \textit{continuity}. The continuity is here to be understood as the mathematical continuity called \textit{microcontinuity}, which requires that infinitely small changes in the input always result in infinitely small changes in the output.  If there is a function of existence mapping to vectors in our domain ontology, which is (at least locally) continuous, we can talk of a \textit{(partially) continuous FOE}.\\

Micro-continuity of a set of vectors inside a sparsely populated vector space and thereby in functions of existence is a rare and, hence, interesting behavior to be observed. This is, independent of its content, continuity inside such a mathematical structure has an a priori reason to be observed, whether our perception pays special attention to it or not. \\

It turns out, however, that the continuity of FOEs is one of the most important patterns discussed in traditional ontology. One of the key distinctions made by previous authors \parencite{BFO, GFO, Guizzardi05} is the one between endurants and perdurants. Endurants (in some literature called continuants) are \textit{things or objects} that exist throughout time; perdurants, on the other hand, include processes or events that have no continuous existence in time \parencite{Guizzardi05, BFO, GFO}. 
Traditional ontology refers to a specific case of continuity in patterns of existence, which are continuous functions given as $f_e(time)$ and sometimes even $f_e(time,space)$. Intuitively, we define endurance as moving continuously through time and space. This evidently requires that time and Euclidean space coordinates be part of the basis vectors of our perceived world ontology.\\

Endurants can be defined as FOEs that are continuous across time; Perdurants, on the other hand, are not continuous FOEs, and in fact, they do not form a coherent subset in a vector ontology at all. In light of a vector ontology, they seem to be a category united by NOT being a continuous function $f_e(t)$. Interestingly, this shows that endurants do indeed have an a priori special property in our formal ontology, given as the continuity. The choice to create a category purely based on continuity in time (a specific basis vector), as done by foundational ontologists, on the other hand, seems rooted in human perception more so than in a priori meaning. \\

Our perception puts categorical importance on stability and, hence, continuity in space-time, which might indicate the true existence of said space-time continuity as a global law. However, another interpretation, one I find more interesting, could be that the psyche or consciousness itself exists as a space-time-continuous function in this ontological form.\\

\textbf{Convex Regions and Mereology}\\

An important ontological issue is that of parthood relations, which have been extensively studied in foundational ontologies. 

Mereology, the branch of ontology concerned with part-whole relationships, solves many of these questions using a form of Set theory \parencite{Guizzardi05} and connected operations such as unions and intersections. I propose using convex regions in the vector ontology to model these relationships or, in more mathematically sound terms, convex subsets or subspaces.

\begin{tcolorbox}[colback=blue!5!white, colframe=blue!75!black, title=Convex subspaces]

Let $V$ be a vector space over the field $\mathbb{R}$, and let $C \subseteq V$. The set $C$ is called \emph{convex} if for all $x,y \in C$ and for every $\lambda \in [0,1]$, the linear combination 
\[
\lambda x + (1-\lambda)y 
\]
also belongs to $C$, i.e.,
\[
\forall\, x,y \in C,\; \forall\, \lambda \in [0,1]:\quad \lambda x + (1-\lambda)y \in C.
\]

Intuitively, this means that a convex subspace has the following property. For any two points in the region, all points on the line connecting them are also in the region, that is the line doesn't cross the regions boundary. On the contrary, a concave region or set has at least two points, which, when connected, cross the boundaries and have parts of the line outside of the region. This mathematical property maps elegantly onto our intuitive understanding of parts and wholes - if an entity A is part of entity B, then A occupies some subset of the "region" defined by B in our quality dimensions.

    \centering
    \includegraphics[width=0.55\linewidth]{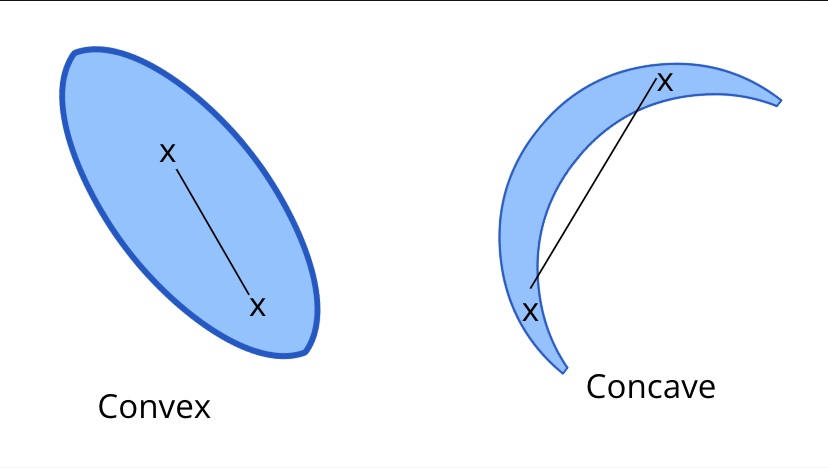}

\end{tcolorbox}

Let's consider the example of a car engine as part of a car. In a vector ontology with spatial dimensions, the engine occupies a convex region that is entirely contained within the larger convex region occupied by the car. However, vector ontologies are powerful because they allow us to consider non-spatial quality dimensions. For instance, in dimensions representing functional properties, the engine's functional characteristics (power\_generated, gas\_consumption, etc.) form a convex region that must be compatible with and contained within the car's overall functional requirements described in the same vector ontology.
Accordingly, I argue that the distinctions in previous work on functional vs. spatial parthood and other forms are not structurally different but rather occur in a subspace with different basis vectors, creating a more cohesive and unified theory for mereological questions.

This formalized definition allows for additional investigation of parthood and its meaning. For example, one might measure the distance between different parts and the center of the overarching structure (e.g., the distance from the engine to the center of the car) to define a measure of \textit{centrality} to, for example, the car's functionality. Lastly, I suspect this definition of parthood will naturally cluster parts that form functional and structural units or subsystems.

The relationship between clusters and convex regions also helps explain why certain parthood relations might feel more "natural" than others. When we observe dense clusters of vectors in our ontological space, separated by relatively empty regions, these often correspond to our intuitive understanding of natural parts or wholes. This provides a formal explanation for why we tend to recognize certain collections of properties as constituting coherent objects or parts while other possible groupings feel arbitrary or artificial.\\

To conclude, I propose that convex regions in vector spaces with a well-chosen basis can potentially capture mereology in a single structural form. This means that parthood is a special type of existence function that determines the convexity of sets, as well as their inclusion and intersection across specific basis vectors.\\

\textbf{Linear dependence and Correlation}\\

Basis vectors in a vector space should be linearly independent. Clearly, that is not the case for most of the ontological universals, such as adjectives that we prescribe to the world. And this creates very obvious problems. Let us have a look at an example.

\begin{tcolorbox}[colback=blue!5!white, colframe=blue!75!black, title=Example of linear dependence in $V_{ont}$]

One considers the colors red, green, and yellow. Let's place them in a vector ontology, which has had great success in display technology with the base vectors of $\{r, g, b\}$ red, green, and blue light. It becomes evident that in such vector space, the "yellowness" of a color is a linear combination of r and g.
    \centering
    \includegraphics[width=0.6\linewidth]{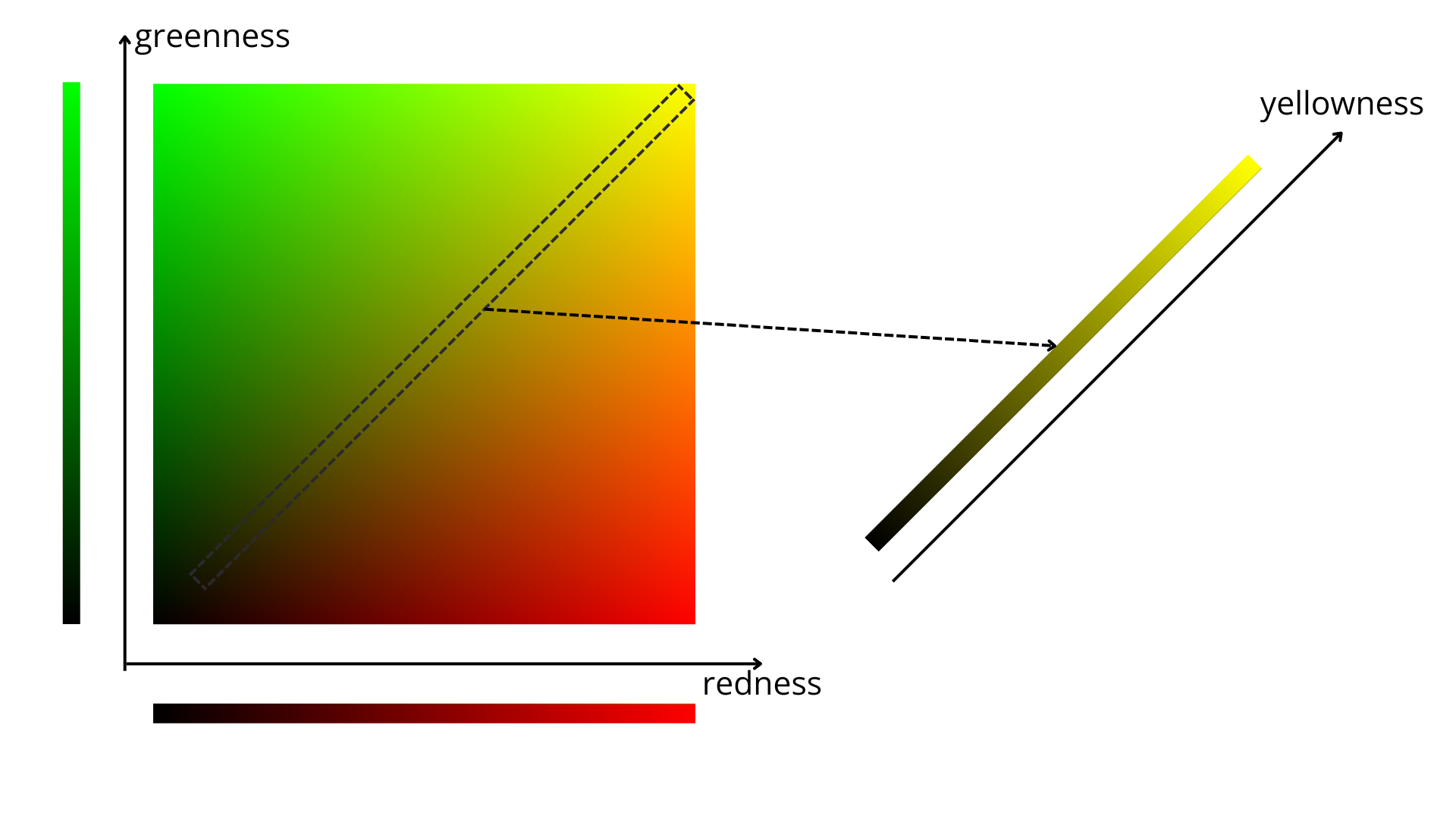}

\end{tcolorbox}
\begin{tcolorbox}[colback=blue!5!white, colframe=blue!75!black]
As such, yellow in this conceptualization can not be used to construct colors together with only 2 of the other vectors. We require all three (r,g,b) to create all possible colors. It is the same phenomenon (although in subtractive rather than additive coloring) that we have experienced as a child when trying to mix green using yellow and blue. Since blue is a linear combination of cyan and magenta, it can not be used with yellow to create true green. The color wheel, as we used to know it with the colors yellow, blue, and red, utilizes linearly dependent basis vectors, which is why it fails to mix all colors.\\
\\
Interestingly, the above shows that we sometimes understand linearly dependent vectors as a separate concept (such as yellow and blue in this case). Importantly, which set of linearly independent basis vectors we choose specifically inside the 3d space seems arbitrary. For example the CMY color space is just a version of the same 3d color space as rgb with the coordinate systems originating in the opposite corner, see below.\\

    \centering
    \includegraphics[width=1\linewidth]{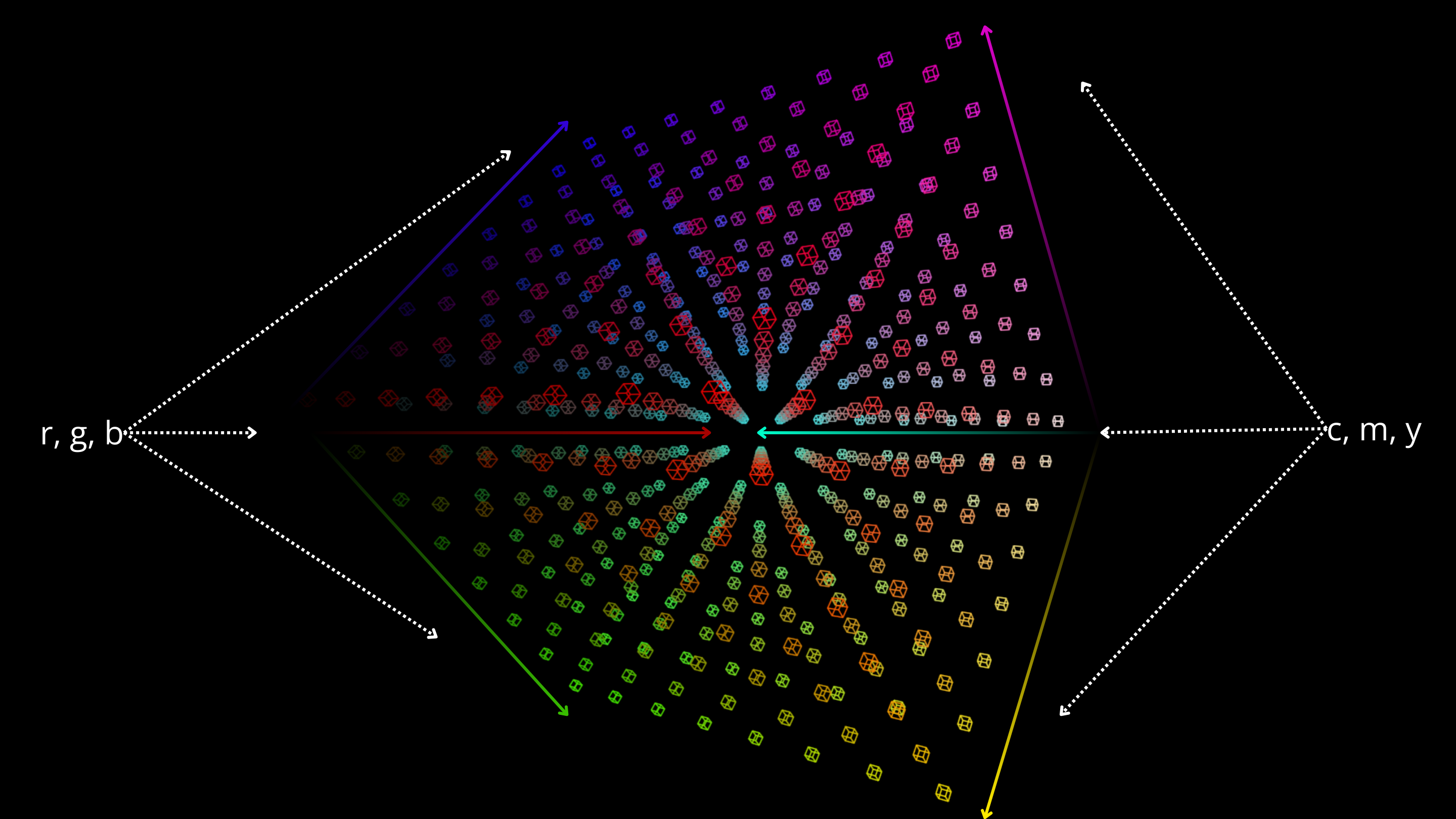}
\end{tcolorbox}

We can conclude that linear dependency in vector ontology causes the same perceived problems (lack of color mixing ability) as the vector ontology would predict, namely that they are not a set of valid basis vectors for the complete vector space and hence struggle to span the space of possible existent vectors (all colors). This is what I argue our perception calls causation. That is, the reason a color space is not sufficiently defined lies in the fact that adding more yellow light is equivalent to adding more green and red light. Hence, the addition of yellow \textit{causes} the addition of green and red.

This directly leads us to the next point, which is that a vector space fulfills the additive and scalar multiplicative closure. That means any vector resulting in the addition of two vectors inside the vector space is also part of the vector space and for scalar multiplication identically.
This would imply that in a domain of colored shapes, for example, with the basis, ${number\_of\_edges, r, g, b}$ any n-edged shape of arbitrary color exists. However, as discussed in previous sections, ontology is interested in a distinction between what could potentially exist, and what truly does. So, instead of full vector spaces as defined in the mathematical structure itself, we assume that reality exists as a sparse population of the theoretically complete Vector space. We have already talked about existence and nonexistence as the truth value of:

\[
v \in V_{ont}
\]

We can, however, also talk about possible existence defined as: 
\[
\Diamond (v \in V_{\text{ont}})
\]

This intuitively is understood as the theoretical possibility of a blue rectangle in a colored\_shape domain, but no guarantee of its materialization. In that case, we can't know precisely whether it exists or not,  raising a logical interest in the \textit{probability} of a specific vector's existence:
\[
{p}\bigl(v \in V_{\text{ont}}\bigr)
\]

Naturally, it would be helpful if we would be able to learn probabilistic functions of existence, let's call them $f_{pe}$, which map from a specific domain to probabilities of existence for a set of vectors:

\[
{p}\bigl(v \in V_{\text{ont}}\bigr) = f_{pe}(x_i, ...)
\]

I argue that probability functions are a common way to express domain knowledge across the sciences. Whether electron positions, weather forecasts, game theory, or reactions to cancer treatment, functions describing probabilistic outcomes are everywhere. In the same way that linear dependence is perceived as causation, I propose that what is intuitively perceived as correlation is, in fact, a learned function of probabilistic existence. That is, we perceive a blue rectangle to be more probable to exist than a 1743-edged scribbled shape.\\

In conclusion, I propose to define causation as a strict linear dependence between vectors representing concepts. Similarly, I propose to understand Correlation as a probability function of existence. This also makes it evident that causation has a drastically different meaning than correlation and can not be investigated with the same methods. Correlations can be learned through pattern detection; however, causation requires more sophisticated and formal methods to identify linear dependence.\\

\textbf{Similarity and Navigation}\\

How we perceive similarity is a question that \cite{Gardenfors} has already linked to vector spaces. The intuition in Gardenfors's findings is that we use a Minkowskian distance metric in a property space to determine the similarity of things, where a smaller distance represents a larger similarity.\\

Minkowskian distance metric (for any r) is defined as:
\[
d(x,y) = \left( \sum_{i=1}^{n} \left| x_i - y_i \right|^r \right)^{\frac{1}{r}},
\]
which computes the distance between the vectors
\[
x = (x_1, x_2, \dots, x_n) \quad \text{and} \quad y = (y_1, y_2, \dots, y_n).
\]

Often we use the special case of $r=2$ which is called the euclidean distance.
\[
d(x,y) = \left( \sum_{i=1}^{n} \left| x_i - y_i \right|^2 \right)^{\frac{1}{2}}.
\]

The correct distance metric to be used in vector spaces is a highly debated topic across the literature of various fields. Much of it applies to vector ontologies the same way it does in any other vector space. However, in our specific ontological case, there is a new notion that yields the  most powerful utility of vector ontologies: reconstruction, extrapolation, and navigability.

\textbf{Reconstruction}\\

An important axiom of vector spaces is the closure. The closure implies that we can reconstruct any vector inside the vector space using a sum of another vector and a serious of scaled basis vectors. This isn't particularly special in an arbitrary vector space and hasn't gotten much attention.

\begin{tcolorbox}[colback=blue!5!white, colframe=blue!75!black, title=Closure in Vector spaces]
    
Let \(V\) be a vector space over a field \(F\) with a basis
\[
\{b_1, b_2, \dots, b_n\}.
\]
Then, by the closure properties of \(V\) (i.e., closure under addition and scalar multiplication), we have:
\[
\forall\, a_1, a_2, \dots, a_n \in F,\quad \sum_{i=1}^{n} a_i\, b_i \in V.
\]
Moreover, every vector \(v \in V\) can be uniquely expressed as a linear combination of the basis vectors:
\[
\forall\, v \in V,\quad \exists! \, a_1, a_2, \dots, a_n \in F \quad \text{such that} \quad v = \sum_{i=1}^{n} a_i\, b_i.
\]
\end{tcolorbox}

This is  because close to all literature on vector spaces has been concerned with uninterpretable feature spaces. Mainly originating from machine learning, the field has been investigating learned features of neural networks, which are mostly meaningless. In a vector ontology, however, every dimension represents an ontologically meaningful dimension. This means that if we understand the concept represented by the quality dimension (e.g., the sweetness of an apple), we have predictive power as to what a change in it represents. This allows us to perform reconstruction and extrapolation since we can do global inference based on local observation. In other words, if we taste $apple_1$, we can extrapolate or reconstruct how $apple_2$ tastes, given we know:
\[
apple_2 = apple_1 + 0.5*[sweetness]
\]

In this scenario, we describe $apple_2$ as an addition of a scaled basis vector (adding sweetness) and a reference vector ($apple_1$). Similarly, a third apple might be given as:
\[
apple_3 = apple_1 + 0.5*[sweetness] + 0.2*[redness]
\]

I propose a generalization of this, which gives a sort of reconstruction distance $D_{reconstruction}$, which is computed using a reconstruction path $P_{reconstruction}$ defined as:

\[
P_{reconstrcution} = v_{origin} + \sum{a*x_i}
\]
Where $v_{origin}$ acts as a reference vector and $x_i$ represent the basis vectors of the vector ontology. Accordingly, we can compute the reconstruction distance as: 
\[
D_{reconstruction} = len(P_{construction})
\]
Where the length of the reconstruction path quantifies the Reconstruction distance. Much evidence, I argue, suggests that our fundamental perceptions of relations are rooted in this reconstruction path and its length. 

\begin{tcolorbox}[colback=blue!5!white, colframe=blue!75!black, title=Reconstruction paths in human perception]

We seem to have no problem intuitively understanding the relation between two objects of the same shape but significantly different sizes. A shape, on the other hand, which has a couple more edges and a slightly different size, seems less similar or less related. You can test that in the graphic below, which clearly shows that the shape in the middle is spatially closer to the small triangle than the large triangle. This is in both Euclidean and Manhattan distances. However, we perceive more relation and similarity between the two triangles than the small triangle and the pentagon. Our perception hence perceives a reconstruction path with fewer steps or corners as shorter. 

    \centering
    \includegraphics[width=1\linewidth]{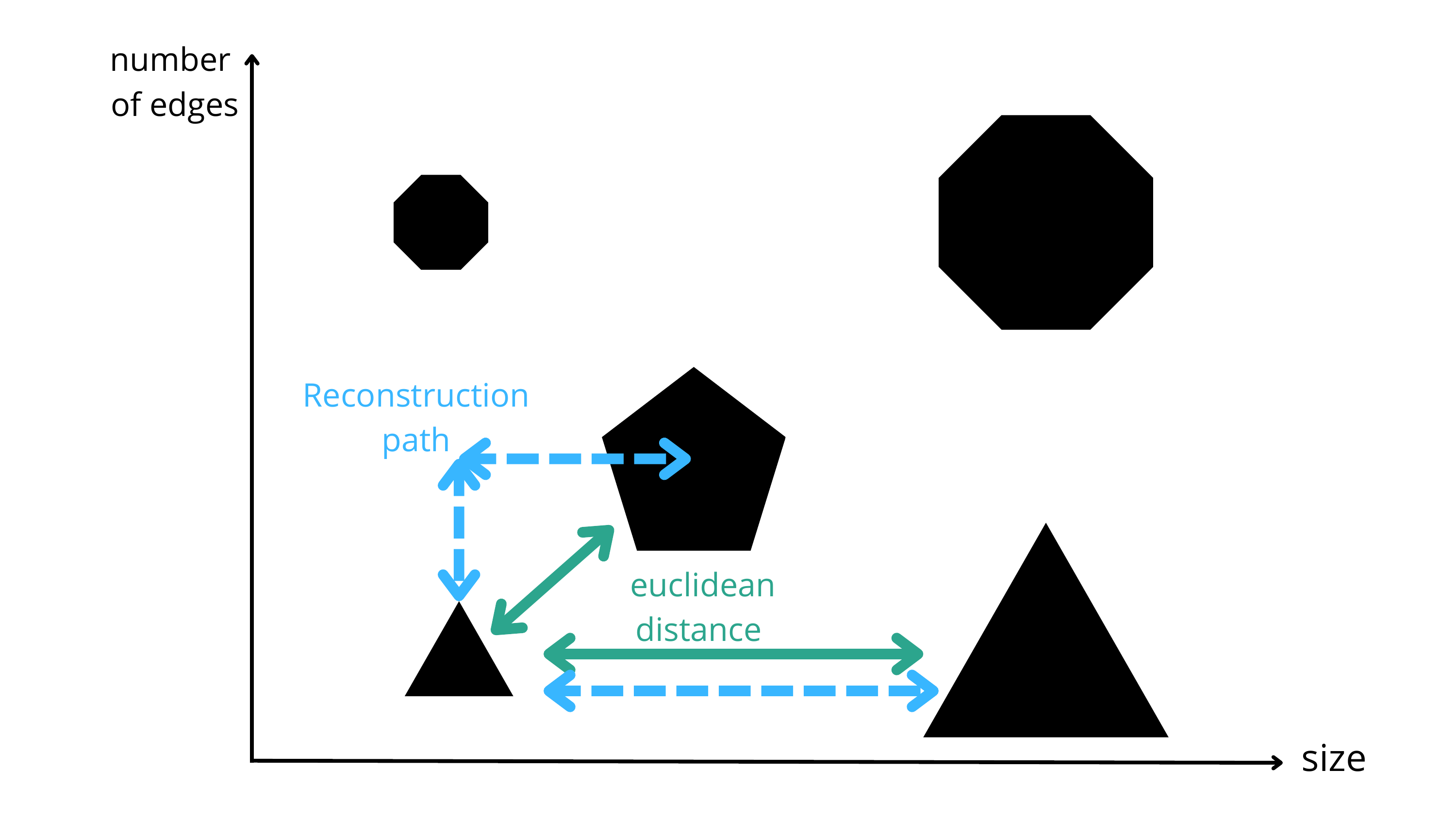}
\end{tcolorbox}

We tend to perceive similarity as a mix of dimensions that are identical and some that are different between two objects. For example, two cars of the same type (e.g., van) but different brands and colors. This makes it evident that the reconstruction path plays a critical role in human perception. It also offers a framework to explain why metaphors are so powerful. We can, for example, learn the dynamics shown in the movement of atoms by comparing it to the movement of planets. This is, we can reconstruct the laws of atoms as a much smaller version of the planetary motion, where the gravitational force is replaced with the electronic force. This means we can express the following equation:
\[
Movement_{atoms} = Movement_{planets} -0.999[size] -1[gravitational\_force] + 1[electric\_force]
\]
Both the Euclidean and Manhattan distance between $Movement_{atoms}$  and $Movement_{planets}$  in this equation would be very large due to the extreme difference in the size of planets compared to atoms. The reconstruction path, on the other hand,  is very short (length 3), which clearly shows the reliance on Reconstruction distance for the comparison of related objects. Knowing the reconstruction path, we can easily grasp the movement of electrons around protons by copying most of our context from planetary motion (the reference vector). As such, the metaphor translates using shifts in a limited set of dimensions which are different between the two cases. The content transferred or the metaphors payload, on the other hand, are those dimensions that are constant, such as those describing elliptic trajectories and centrifugal forces, creating a "shared context."  This type of transfer learning has also been suggested by Gardenfors \parencite{Gardenfors}  as a fundamental dynamic of how we learn new concepts using our knowledge of other ones.\\

This showcases how vector ontologies encode context and relations. Paths connecting point A (origin vectors) to point B (the target vector) while only moving in a restricted set of basis vectors act as a sort of reconstruction or navigation instructions, which tell us how to manipulate the origin vector in order to find the target. As such paths between objects represent relations we perceive between them as a set of movements through quality dimensions. Since the basis is interpretable this makes vector ontologies directly navigable. That is, in an ontology of books, we can find the scientific version of our favorite comic or the grown-up version of Harry Potter if $scientificness$ and $age\_of\_audience$ are expressible as combinations of the basis vectors in the book ontology. It is this navigability which I expect to be an incredibly valuable property of vector ontologies in search tasks.\\

The introduction of Paths between objects as meaningful relationships also draws a link to Graph theory. Although not completely homomorphic, the vector space can be seen as a folded graph in which predicate types are encoded as paths through a limited set of dimensions. For example, the $is\_small\_model\_of$ relation is described as a movement in the size dimension while keeping most other dimensions stable.

\section{Humans and vector ontologies}

The relation between human perception and Vector ontologies is largely covered in the sections above relating geometric phenomena to ontological perception. However, Gardenfors has already demonstrated more direct evidence that humans use vector ontologies in many different scenarios \parencite{Gardenfors}, including color perception, taste, height, and other properties. He roots this theory largely in the cognitive sciences and experimental methods, which clearly support his hypothesis of convex regions displaying types and similarity as a function of distance. To keep things concise, I refer to his work for further reading if one wishes to dive deeper into the empirical evidence for humans' use of vector ontologies.

\section{Suggested Extensions to Vector Ontologies}

It's likely that to capture some ontological phenomena of reality, an extension from vector ontologies to tensor ontologies with more complex topologies will be required. This is a logical step and would impact the above-elaborated concepts as the logical extensions of tensor mathematics. I believe these extensions to be sufficiently logical and deductible for mathematically educated readers to skip elaboration at this point and move them to future work.

\section{Resolution and the Cost of Vector Ontology}

Analysis of vector ontology is notoriously computationally intensive. This is due to the curse of dimensionality, which renders computations across many dimensions exponentially costly. Hence, we will need to choose basis vectors wisely and think about the required resolution for a given use case.

As mentioned above, I suspect that individual domains (subsets/subspaces) have a finite set of basis vectors, which, similar to the Fourier transform, can be sorted by importance. Hence, I suspect an asymptotic graph where, by adding dimensions, we asymptotically approach true (in the sense of a priori true in our structure) representation but never reach it. If the most impactful basis vectors can be sucessfully identified, however, I suspect that a finite and relatively manageable number of dimensions will suffice (tens to hundreds).

This, will clearly depend on the degree to which the domain is in fact a domain defined by shared basis vectors.\\

I argue that both the computational complexity and the asymptotic behavior are strengths of this approach. Computation in vector ontologies, while exponentially expensive, also gives us exponentially rich returns. This is because it yields new information for all of the vectors across all dimensions. As such, we can capitalize on the full compute. Intuitively, my argument is that in this case, adding dimensions gives us an exponential increase in knowledge/understanding. Hence, the compute is better spent than when doubling the context window of LLMs while quadrupling the compute, where the basis value (context length) scales linearly and the cost scales quadratically. 
Secondly, and more importantly, this structure gives us an asymptotic approach to reality, which acknowledges its insufficiency in that it can never reach it. This approach also clearly gives us an opportunity to increase our success with increased work (computation).

\section{Expected Criticism}

The goal of this paper is to spark a discussion among researchers in ontology, computer science, and cognitive sciences. While I am looking forward to hearing new viewpoints on the issue, I already predict several arguments to be raised. To speed up the scientific discussion, I will, hence, already provide an initial rebuttal of those points in the following section.

\subsection{Formal vs. Foundational ontology}

Some might disagree with the claim that vector ontologies are truly formal, contrary to the categorical ontologies developed by \parencite{Husserl_2, BFO, GFO, Guizzardi05} , and others. Specifically, they might bring forward one of the following two arguments.\\

\textbf{1. Current ontologies are already formal in the sense that they are independent of any domain. }\\

I agree with the notion that this makes them formal in the broadest definition of the \textit{word} "formal". However, "formal", as used by Husserl in philosophy and in formal mathematics, has a much more rigorous definition. This definition can not be satisfied through recursive abstraction (see section 2). The term "foundational" encapsulates the notion of domain independence much more accurately than "formal" and hence should be used for recursively abstracted categorical ontologies.\\

\textbf{2. Vector ontologies introduce perceptive bias the same way categorical ontologies do by choosing basis vectors.}\\

I fully agree with this statement; however, there is an important distinction to be made. Vector ontologies define complex rules in their axioms \textit{before} choosing any basis vectors. As such, the \textit{structure} of the ontology or its \textit{form} exists before introducing a basis and, hence, perceptive bias. Once we choose basis vectors, the axioms of the vector space create complex interactions inside the ontology. These interactions, however, arise from the predefined axioms, not the chosen basis. As such, we can check whether the interactions that arise align with our ontological perception, acting as a sort of reality check on the chosen basis. 
This is drastically different from categorical ontologies, which do not define any structure before introducing perceptive bias and hence have nothing to check perception against.

\subsection{Oversimplification}

Some might argue that expressing every ontological construct inside a vector space oversimplifies the true complexity of reality.  Although I understand this concern intuitively, I think it lacks full grasp of both the alternative ontological structures in use today and the complexity of vector spaces. 

The foundational ontologies built by current literature describe categories such as objects, properties, modes, endurants, and events \parencite{Husserl_2, BFO, GFO, Guizzardi05} and relationships as Subject predicate Object triples. Mathematically, categorizations instantiate set theory and subject predicate object triples instantiate graphs, which are both much simpler mathematical structures than vector spaces (or tensors). Accordingly, I argue that claiming vector ontologies to be an oversimplification is incoherent with the current categorical and relationship based ontologies, which instantiate the much simpler structures and yet are believed to capture reality.

\subsection{Lack of Identity and types}

Gardenfors' preposition of Conceptual spaces has been criticized for lack of \textit{identity} by \cite{Guizzardi15}, which makes it logical to expect similar criticism for vector ontologies. The argument broad forward by \cite{Guizzardi15}, when applied to vector ontologies rather than conceptual spaces, goes roughly like this:\\

If an object is defined as a vector in a high dimensional space describing its qualities, how do we handle qualitative changes of the same object over time. A Vector ontology assumes a Leibnizian principle of identity, which is based on the identity of indiscernibles. This means that there can not be two distinct objects with the exact same properties, and accordingly, there can not be two distinct vectors at the exact same position. If two vectors are at the same position, they represent the same object. This is problematic, as a single object might change over time while keeping its identity. For example, a child becomes an Adult, clearly changing its position in time, space, and properties such as height, weight, strength, and so forth, and accordingly exists as a series of points, not a single one. This creates a problem, as the identity of the person across time can not be described as a constant position or any other form of constant, so \cite{Guizzardi15} argues. Furthermore it would be impossible to describe possible worlds where the same person has for example a different occupation, as there is not way to tie these possible versions of the person to a single position in the vector space. Rather, a person would be represented by a series of points across time, and it becomes unclear how to determine which points belong together as a single person. To solve this issue, Guizzardi suggests going beyond static representations by introducing a “kind” or sortal concept. This additional structure specifies which properties are essential and must remain invariant for an individual to retain its identity, even as other properties change over time. He proposes that such a Concept definition is outside the vector space itself and should be implemented as some \textit{structure} associating \textit{sortals} with individual concepts that can be thought of as a "projection into the conceptual space defining a suitably constrained \textit{set of points }that represent counterparts in different worlds of the same ordinary object" (p.28). As such, the structure defines types of objects (e.g., Person, Dog, ...), and individual concepts represent instantiations of the same. 
 
I argue that this is largely congruent with the \textit{Functions of (possible) existence} as defined in section 4.1.3. A projection as described by Guizzardi when translated into mathematical terms is essentially a linear map or function. Hence, identity as a "projecting structure," as proposed by Guizzardi, mathematically resembles a function of existence, in line with what we have described above, where we learn functions that describe many vectors as their output space to compress many points into a single concept. In line with Guizzardi, I argue that functions providing identity tend to be continuous or even constant in some dimensions, meaning that we expect continuity in order to keep identity which is directly tied to their endurant property (see section 4.1.3). 

This leaves the question of the structure that describes not an individual concept but the type itself, and some might question the ability of a vector ontology to express types. 

\cite{Lopes} made a related argument in his investigation of Convolutional Neural Networks (CNNs) and their ability to learn types or concepts. He argues that vector-based ontological systems eventually define clusters through similarity or closeness in a vector space. This, so he argues, does not clearly distinguish types as it doesn't draw direct boundaries but simply estimates relatedness to other instances of objects. Similarly, \cite{Guizzardi15} argues that vector spaces can't capture the notion of a type since a type incorporates not just instantiations or regions of the object that belong to the type but everything that possibly could belong to the type. Hence,  defining the type by a region of properties becomes a circular definition where the boundary is drawn based on the type, and the type is defined based on the boundary, creating a question of origin.

I argue that this problem does not occur in the vector ontology as defined above. This is because concepts are not regions but rather Functions of existence, which could be describing regions but might describe waves, lines, or anything else. As such, types would similarly not be regions but rather classes of functions. For example, a type \textit{sphere} necessarily takes the general form  $(x+a)^2+(y+b)^2+(z+c)^2 \le r^2$, but any individual sphere instantiates as a concrete instance of the equation with $a, b, c, r$ taking discrete values.

As such types can in fact be described in a vector ontology as general function classes with tunable parameters.

\section{Conclusion}

I clearly demonstrated that what has so far been classified as formal ontology has not been formal in a strict Husserlian sense. I also constructively explained the potential of a truly formal ontology as a more objective and rigid approach to what is, which can design its form to become an asset. Lastly, I clearly show the potential of a specific class of formal ontology, namely vector-ontologies, due to three key reasons. Firstly, they are already in used implicitly by machines and humans and act as a natural ontological interface between the two. Secondly, they can effectively capture existing ontological structures while adding mathematical structure to them, thereby explaining their relation to each other and building a more comprehensive perspective on ontology as a whole. Lastly, their generalized structure, which can hold an arbitrary number of meaningful dimensions, makes them scalable while staying interpretable and, hence, navigable. This navigability, I believe, together with their machine interoperability, will make them orders of magnitude more useful in automated knowledge tasks than traditional ontologies.

\section{Future Work}

As mentioned above, I hope this argument sparks a discussion in philosophy, computer sciences, and related fields as to the utility of formal ontological frameworks. However, since this Paper is theoretical, future work should also empirically test the proposed vector ontological structure. Concretely, I suggest experiments to test the following implicit hypothesis following from this paper:

\begin{enumerate}
    \item We can define Vector ontologies for specific domains that fully encapsulate their ontological structure. 
    \item We can unify the ontological structure for several domains in a single  vector ontology
    \item We can unify all domains in a single vector ontology
    \item ANNs utilize vector ontologies in their internal representations (which we can understand)
    \item We can extract vector ontologies from ANNs
    \item ANNs can understand predefined vector ontologies
    \item We can build interpretable and highly ontological capable artificial knowledge systems using Vector ontologies
\end{enumerate}

\section{Acknowledgments}

This work would not be possible without the support of my colleague Bekk Blando, whose skepticism, rigor, and intellectual input have greatly sharpened this argument and who is collaborating with me on the empirical proofs that we will publish within the next weeks. I am also thankful to  Elisabeth Bühler and Mats van Dalen, who have contributed to the early conceptual developments.

\section{References}
\setlength{\bibitemsep}{10pt}
\printbibliography
\end{document}